\documentclass{article}

\usepackage{microtype}
\usepackage{graphicx}
\usepackage{subfigure}
\usepackage{booktabs} 
\usepackage{xcolor}
\definecolor{linkcolor}{RGB}{0,128,255}
\usepackage[colorlinks=true,allcolors=linkcolor,pageanchor=true,plainpages=false,pdfpagelabels,bookmarks,bookmarksnumbered]{hyperref}

\usepackage[T1]{fontenc}    
\usepackage{nicefrac}       
\usepackage{amsmath,amssymb,amsfonts}
\usepackage{wrapfig}
\usepackage{booktabs}
\usepackage{mathtools}
\usepackage{dsfont}
\usepackage{commands}
\usepackage{natbib}
\usepackage{setspace}
\usepackage{xspace}
\usepackage[nameinlink]{cleveref}
\usepackage{subfigure}
\usepackage{amsthm}

\usepackage{lscape}

\usepackage{todonotes}

\usepackage{algorithm}
\usepackage{algorithmic}
\newtheorem{theorem}{Theorem}
\newtheorem{proposition}[theorem]{Proposition}

\newtheorem{definition}[theorem]{Definition}

\newtheorem{remark}{Remark}

\DeclareMathOperator*{\argmin}{arg\,min}

\usepackage{hyperref}

\usepackage[accepted]{icml2021}

\icmltitlerunning{Estimating Barycenters of Measures in High Dimensions}

\begin{document}

\twocolumn[
\icmltitle{Estimating Barycenters of Measures in High Dimensions}



\icmlsetsymbol{equal}{*}

\begin{icmlauthorlist}
\icmlauthor{Samuel Cohen}{ucl}
\icmlauthor{Michael Arbel}{gat}
\icmlauthor{Marc Peter Deisenroth}{ucl}
\end{icmlauthorlist}

\icmlaffiliation{ucl}{Centre for Artificial Intelligence,  University College London, UK}
\icmlaffiliation{gat}{Gatsby Computational Neuroscience Unit, University College London}

\icmlcorrespondingauthor{Samuel Cohen}{samuel.cohen.19@ucl.ac.uk}

\icmlkeywords{Machine Learning, ICML}

\vskip 0.3in
]



\printAffiliationsAndNotice{} 
\begin{abstract}
Barycenters are  principled summaries of populations of measures. To estimate barycenters, we typically parametrize them as weighted sums of Diracs and optimize their weights and/or locations. This approach, however, does not scale to high dimensions due to the curse of dimensionality. In this paper, we propose a technique for facilitating difficult barycenter estimation problems through a different parametrization of the barycenter by means of a generative model. This turns the barycenter estimation into an optimization problem over model parameters, which sidesteps the curse of dimensionality and allows for incorporating inductive biases directly into the model. We prove local convergence under mild assumptions on the discrepancy, thereby showing that the approach is well-posed. We demonstrate that our method achieves good performance on low-dimensional problems and provide unprecedented results by scaling barycenter estimation effectively to high-dimensional image data. 
\end{abstract}


\section{Introduction}

Barycenters  are principled summaries (averages) of probability measures \citep{journals/siamma/AguehC11}, defined with respect to a similarity metric on the space of measures. They have been used in computer vision \citep{gramfort:hal-01135198}, economics \citep{carlier:hal-00987292}, Bayesian inference \citep{Srivastava2015}, physics \citep{pmlr-v48-peyre16}, and machine learning \citep{dognin2018wasserstein}. 

Computing barycenters has been extensively studied by  \citet{journals/siamma/AguehC11,pmlr-v32-cuturi14,benamou:hal-01096124,NIPS2019_9130}. It is extremely challenging, due to the need to optimize over spaces of measures. Current approaches typically use compactly-supported basis functions, in particular Diracs, to parametrize barycenters and optimize their weights and\slash or locations \citep{pmlr-v32-cuturi14,NIPS2019_9130}. The strictly local property of these functions requires an exponentially increasing number of basis functions as the dimensionality of their domain increases. As a result of this `curse of dimensionality', these methods are typically restricted to low-dimensional problems ($\mathbb{R}^{\leq 3}$).  From a theoretical standpoint, \citet{altschuler} indeed highlights the NP-hardness of computing Wasserstein barycenters of measures, and hence the dimensionality curse. As a result, algorithms that do not incorporate structure (and leverage the low-dimensional structure in high dimensions) are doomed in high dimensions.  Concurrent work by \citet{Shen2020SinkhornBV} takes a global approach to computing Sinkhorn barycenters and exploits a form of functional gradient descent to scale better with respect to dimensions than local methods.  This approach, however, is limited to averaging under the Sinkhorn geometry, and was only used in synthetic settings.

In this paper, we introduce a practical algorithm for estimating barycenters that can be applied to high-dimensional settings. The key idea is to use a different parametrization of the barycenter by means of a generative model, turning the optimization over measures into a more tractable optimization over  parameters of the generative model. 
For instance, when learning a barycenter of measures  on image space, we parametrize a CNN generating images, instead of parametrizing individual images constituting the barycenter.

Importantly, our approach allows to enforce a \emph{global structure} by treating the barycenter as a parametric model instead of a collection of point masses. It also introduces inductive biases to the model that can reach accurate solutions faster. The combination of global structure and inductive biases in the generator allows us to apply our algorithm to barycentric problems at unprecedented scales in terms of dimensions and support (e.g., in image space $\mathbb{R}^{\text{width}\times \text{height}\times \text{channels}}$). We also demonstrate that our approach leverages the problem structure to obtain additional speedups by incorporating inductive biases.


We also study convergence properties of our proposed algorithm to stationary points for general choices of  discrepancies. In particular, we show that local convergence holds for all discrepancies that are either Lipschitz smooth or weakly-convex and Lipschitz continuous, which includes Sinkhorn as proved in \citet{10.5555/3327757.3327812} and MMD (with deep kernel) as proved in this paper. 
We apply our algorithm to both traditional low-dimensional experiments (e.g., nested ellipses in $\mathbb{R}^2$ \citep{pmlr-v32-cuturi14}), and previously untackled high-dimensional experiments (e.g., on image datasets in $\mathbb{R}^{>10,000}$) for different choices of discrepancies, namely MMD, optimized MMD, and Sinkhorn. To the best of our knowledge, this is the first approach for estimating barycenters that is applied to non-toy, non-synthetic data in high dimensions.

\section{Barycenters of Measures}
\label{sec:background}
We consider the problem of computing barycenters of probability measures defined on a subset $\mathcal{X}$ of $\mathbb{R}^d$.  We denote by $\mathcal{M}_{1}^{+}(\mathcal{X})$ the set of such measures on $\mathcal{X}$ and define the probability simplex $\Delta_P := \{\v{\beta} \in \mathbb{R}^{P} : \sum_{p=1}^{P}\beta_p=1, \beta_p\geq 0\}$.
Following \cite{journals/siamma/AguehC11}, the barycenter of $P$ probability measures $\mu_{1},\ldots,\mu_{P}\in \mathcal{M}_{1}^{+}(\mathcal{X})$  weighted by a vector $\v{\beta} \in \Delta_P$ can be expressed as the measure $\mu^{\star}$ solving 
\begin{align}
    \mu^{\star}=\argmin_{\mu \in \mathcal{M}_{1}^{+}(\mathcal{X})} \sum_{p=1}^{P}\beta_{p}D(\mu, \mu_{p}),
    \label{eq:bary}
\end{align}
where $D: \mathcal{M}_{1}^{+}(\mathcal{X}) \times \mathcal{M}_{1}^{+}(\mathcal{X})\to \mathbb{R}^+$ is a discrepancy between measures .  Depending on the choice of $D$, barycenters have significantly different properties. We discuss two families of barycenters obtained when using the Wasserstein distance and the maximum mean discrepancy (MMD) as discrepancy $D$ mainly based on the works of \citet{DBLP:conf/birthday/BottouALO17, journals/siamma/AguehC11, article_anderes}.
The characterization of barycentric properties will be useful to interpret results in the experiments section.

\subsection{Wasserstein Barycenters}

The \emph{$k$-Wasserstein distance} between two measures $\mu_{x}, \mu_{y}\in\mathcal{M}_{1}^{+}(\mathcal{X})$ is defined as \citep{villani}
\begin{align}
    &\hspace{-2mm}\mathcal{W}_{k}(\mu_{x},\mu_{y})= \min_{\pi \in U(\mu_{x},\mu_{y})} \Big(\int_{\c{X}\times \c{X}}d^{k}(\v{x},\v{y})d\pi(\v{x},\v{y})\Big)^{\frac{1}{k}},\label{eq:wass}
\end{align}
where $d:\mathcal{X}\times \mathcal{X}\to \mathbb{R}$ is a distance representing the cost of transporting a unit of mass from $\v{x} \in \mathcal{X}$ to $\v{y} \in \mathcal{X}$, and $U(\mu_{x},\mu_{y})$ is the set of joint distributions with marginals $\mu_x,\mu_y$. Intuitively,    $\mathcal{W}_k$ in~\eqref{eq:wass} corresponds to  the minimal expected cost of transporting mass from $\mu_{x}$ to $\mu_{y}$ according to an optimal \textit{plan} $\pi\in U(\mu_{x},\mu_{y})$. 

In general, computing the Wasserstein barycenter requires evaluating ~\eqref{eq:wass} several times, which is computationally challenging. Recent advances provide algorithms to solve~\eqref{eq:wass} approximately with a lower computational cost. \citet{Cuturi:2013:SDL:2999792.2999868} proposed to solve a regularized version of~\eqref{eq:wass} by adding a small (relative) \emph{entropic} term for regularization purposes, leading to a smooth convex objective 
\begin{align}
    \mathcal{W}^{k}_{k,\epsilon}(\mu_{x},\mu_{y})\!=\!\min_{\substack{\pi\in U}} \int\! d^{k}(\v{x},\v{y})d\pi(\v{x},\v{y})\!+\!\epsilon \f{KL}(\pi||\mu_x, \mu_y)
    \label{eq:entwass}
\end{align}
for which optimization scales considerably better. Here, $\epsilon\geq 0$ controls the regularization. For simplicity, we refer to $\mathcal{W}_{\epsilon}$ as the entropic-regularized Wasserstein. 

The objective in \eqref{eq:entwass} is biased  as in general $\c{W}_{\epsilon}(\mu,\mu)\neq 0$ 
 \citep{pmlr-v84-genevay18a}. Thus, \eqref{eq:entwass} does not define a distance.  Also, the bias can lead to possibly wrong minima during optimization \citep{g.2018the}. To alleviate this issue, \citet{pmlr-v89-genevay19a} introduced the Sinkhorn divergence 
\begin{align}
\c{S}\c{W}_{\epsilon}=2\c{W}_{\epsilon}(\mu_{x},\mu_{y})-\c{W}_{\epsilon}(\mu_{x},\mu_{x})-\c{W}_{\epsilon}(\mu_{y},\mu_{y}),
\label{eq:sink}
\end{align}
which removes that bias. Equation \eqref{eq:sink} is symmetric, non-negative, and unbiased while still approximating the Wasserstein distance for $\epsilon\to 0$. Hence, the Wasserstein barycenter can be in principle estimated using the Sinkhorn divergence instead of the less tractable Wasserstein distance~\citep{NIPS2019_9130}. We follow this approach in the paper.

\paragraph{Characterization of Wasserstein Barycenters}
It is well-known that Wasserstein barycenters of measures have interpolation properties. 
We state formally the (known) result \citep{journals/siamma/AguehC11}, which will be useful in understanding the behavior of Wasserstein barycenters in high dimensions later. For completeness, we also provide a (new) proof in the Appendix of this known result.

\begin{proposition} \label{the:wass} (2-Wasserstein Barycenter): When the discrepancy between measures is $D=\c{W}_{2}^{2}$, and $d$ is the Euclidean $L^2$ norm, the barycenter $\mu^{\star}$ of measures $\mu_{1}, ...,\mu_{P}\in \mathcal{M}_{1}^{+}(\mathcal{X})$ with weights $\v{\beta} \in \Delta_P$ is 
\begin{align}
\m{Y}\sim \mu^{\star} \iff   \m{Y} = T(\m{X}), \quad \m{X}\sim \pi^\star,
\end{align}
where $\pi^\star$ is a multi-marginal transport plan (see Appendix for more details), and $T(\m{X}) = \frac{1}{P}\sum_{p=1}^P\v{x}_p$.
\end{proposition}

\begin{figure}
\centering
\subfigure[Wasserstein]{
  \includegraphics[width=0.43\hsize]{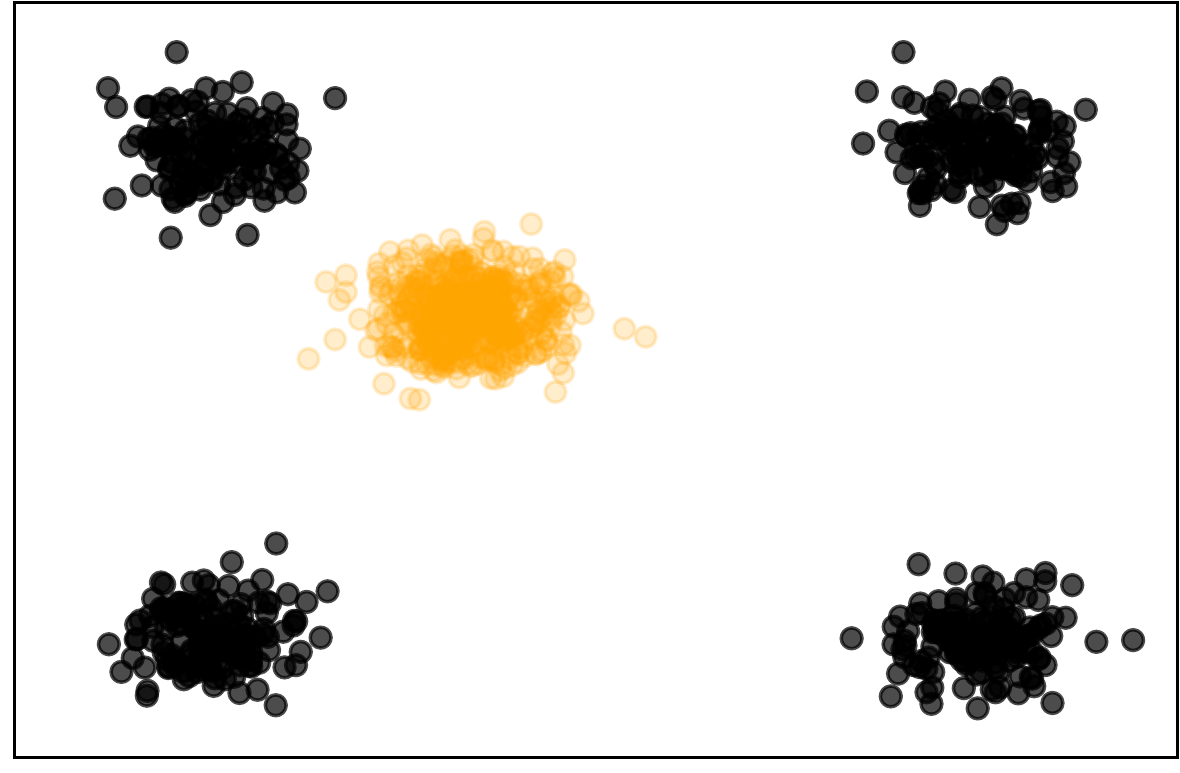}
  \label{fig:wasse}
  }
  \subfigure[MMD]{
  \includegraphics[width=0.43\hsize]{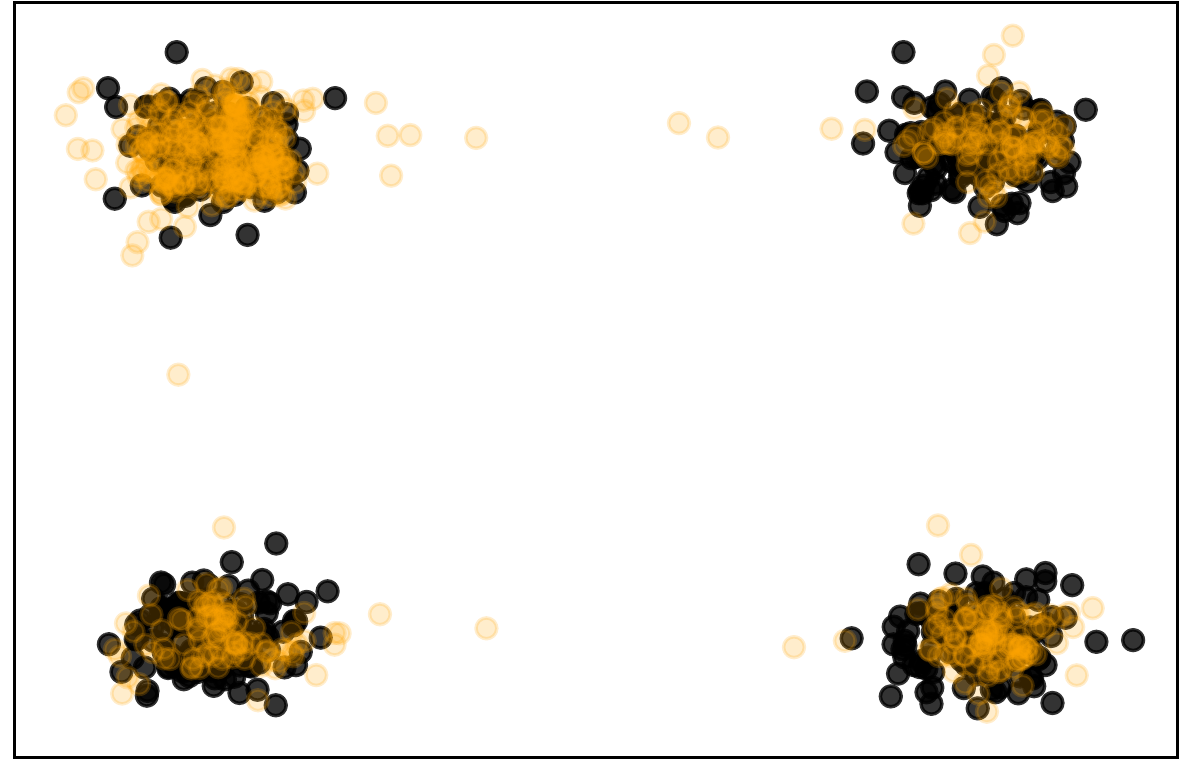}
  \label{fig:mmd}
  }
  
\caption{Barycenter (orange) of four Gaussians (black) with respect  to \subref{fig:wasse} $\c{W}_\epsilon$;  \subref{fig:mmd} MMD. Top-left Gaussian has three times the weight of the others: $\beta = [3/6,1/6,1/6,1/6]$.}
\label{fig:behaviors}
\end{figure}

This means that a sample $\m{Y}$ from the barycenter distribution can be obtained by computing the Euclidean barycenter of samples $\m{X} = (\v{x}_1, ..., \v{x}_p)$ from a joint optimal coupling $\pi^\star$ of $\mu_1,\ldots,\mu_P$, i.e., $\m{Y} = T(\m{X})$.
As an illustration of Proposition \ref{the:wass}, Figure \ref{fig:wasse} shows that the 2-Wasserstein barycenter of four isotropic Gaussians located on the corners of a square indeed displaces the mass proportionally to the weights toward the mode with the highest weight (top left).  

\subsection{(Scaled) Maximum Mean Discrepancy Barycenters}
\label{sec:mmd}

The maximum mean discrepancy \citep{Gretton:2005:MSD:2101372.2101382} 
\begin{align}
   &\mathrm{MMD}(\mu_{x},\mu_{y})^2 :=\mathbb{E}_{\v{x},\v{x}'\sim \mu_x}[k(\v{x},\v{x}')]\nonumber \\&\quad + \mathbb{E}_{\v{y},\v{y}'\sim \mu_y}[k(\v{y},\v{y}')] - 2 \mathbb{E}_{\v{x}\sim \mu_x, \v{y}\sim\mu_y}[k(\v{x},\v{y})] \label{eq:mmd_explicit}
\end{align}
is a discrepancy between probability distributions and relies on a positive definite kernel $k:\mathcal{X}\times \mathcal{X}\rightarrow \mathbb{R}$ 
as a measure of similarity between pairwise samples.
The first two terms in \eqref{eq:mmd_explicit} compute the average similarity within each of $\mu_x$ and $\mu_y$  while the last term computes the average similarity between samples from $\mu_x$ and $\mu_y$.  Unlike the Wasserstein, estimating the MMD using samples from $\mu_x$ and $\mu_y$ is straightforward \citep{Gretton:2005:MSD:2101372.2101382}.

\paragraph{Characterizing the MMD Barycenter}
\begin{proposition} \label{the:mmd} (MMD Barycenter): If $D=\text{MMD}^{2}$, the barycenter of measures $\mu_{1},..., \mu_{P}\in \mathcal{M}_{1}^{+}(\mathcal{X})$ with weights $\v{\beta} \in \Delta_P$ is the mixture of measures 
\begin{align}
\mu^{\star}:=\sum_{p=1}^{P}\beta_{p}\mu_{p} \in \mathcal{M}_{1}^{+}(\mathcal{X}).
\end{align}
Proof in Appendix \ref{sec:proofmmd}.
\end{proposition}
 Proposition \ref{the:mmd} can be seen as a direct extension of  results describing the geodesic structure induced by the MMD (Th. 5.3 in  \citet{DBLP:conf/birthday/BottouALO17}). It also suggests a basic generative process for sampling from MMD barycenters: (i) generate a draw $z \sim \text{Categorical}_{P}(\v{\beta})$; (ii) sample from  measure $\mu_{z}$. Samples from the MMD barycenter (following this procedure) are shown in Figure \ref{fig:mmd}.

 \paragraph{Scaled MMD}

Using MMD with a fixed kernel $k$ is ineffective, e.g., when training generative models on datasets of  images, as the training signal may be small  \citep{NIPS2017_6815}. To alleviate this, deep kernels $k_{f_\psi}(x,y)=k(f_\psi(x),f_\psi(y))$ can be used \citep{Calandra2016, Wilson2016}. When the feature $f_{\psi}$ is fixed,  Proposition \ref{the:mmd} applies and the barycenter is still a mixture of measures. However, learning the feature along with the generator in an adversarial fashion has proven to be more effective \citep{NIPS2018_7904,binkowski2018demystifying,NIPS2017_6815}, allowing  the gradient signal to increase at locations where measures differ. In this case, $D$ is of the form
\begin{align}
   \text{SMMD}^2(\mathbb{P}_{\theta},\mathbb{P}) := \sup_{f_{\psi}\in \mathcal{E}}\lambda(\psi)\text{MMD}^2_{\psi}(\mathbb{P}_{\theta},\mathbb{P}),
\end{align}
where $\lambda(\psi)$ is a scaling function that acts as a regularizer. We assume that all $f_\psi \in \mathcal{E}$ are continuously parametrized by $\psi \in \Psi$ with $\Psi$ compact. Because the kernel changes during training, Proposition \ref{the:mmd} no longer applies, and as a result the barycenters of SMMD might have a different from. Nevertheless, in Section \ref{sec:experiments}, we provide empirical evidence that it retains properties of the MMD barycenter.

\subsection{Related Work on Barycentric Computations}

Most previous approaches to computing barycenters  can be categorized into fixed \citep{pmlr-v32-cuturi14,NIPS2017_6858,NIPS2018_8274} and free \citep{pmlr-v32-cuturi14,pmlr-v80-claici18a,NIPS2019_9130} support. Fixed-support approaches choose a finite set of locations $\v{x}_1,...,\v{x}_N \in \c{X}$, parametrize the barycenter as a weighted sum of Diracs $\mu=\sum_{n=1}^{N}a_{n}\delta_{\v{x}_{n}}$, and optimize \eqref{eq:bary} with respect to weights $a_{n}$. Free-support approaches typically optimize both locations $\v x_{n}$ and weights $a_{n}$ by alternated optimization.

However, these methods hardly scale to high-dimensional problems due to the need to optimize locations $\v{x}_n$. The number of parameters  to optimize scales exponentially with the dimensionality of the space, which makes them inapplicable to high-dimensional problems, such as considering datasets of images (where individual $\v{x}_n$ are images). Indeed, estimating barycenters without enforcing structure is doomed in high dimensions as demonstrated in various theoretical works. For instance, \citet{altschuler} show NP-hardness of the Wasserstein barycenter problem, which  highlights the curse of dimensionality.

As a result of this computational challenge, previous approaches exclusively tackled problems in $\mathbb{R}^{\leq 3}$ \citep{pmlr-v32-cuturi14,benamou:hal-01096124,NIPS2018_8274,pmlr-v80-claici18a,NIPS2019_9130, bonneel:hal-00881872}. 
Concurrent work by \citet{Shen2020SinkhornBV} tackles the problem of estimating Sinkhorn barycenters via functional gradient descent on the push-forward mapping (in a RKHS) of a base measure. The method is therefore tailored to Sinkhorn barycenters, while in this paper we propose a general method that can be used with other choices of $D$. Also, we propose, to the best of our knowledge, the first approach that is demonstrated to work in high-dimensions on non-synthetic data.

\section{Estimating Barycenters Using Generative Models}

In the following, we propose an algorithm for estimating barycenters  between $P$ probability measures with discrepancies including MMD, MMD with optimized kernel \citep{NIPS2018_7904}, $\c{W}_\epsilon$ \citep{Cuturi:2013:SDL:2999792.2999868} and $\c{S}\c{W}_\epsilon$ \citep{pmlr-v84-genevay18a}. 
The key idea behind our algorithm is to parametrize the barycenter using a generative model and thereby turn the optimization over the intractable space of measures into learning model parameters. With this we leverage the fact that high-dimensional data typically lies on significantly lower-dimensional manifolds. This also allows us to incorporate structural inductive biases (e.g., through a CNN), enabling our algorithm to scale to high dimensions. We also prove local convergence for common discrepancies.

\subsection{Algorithm}
\label{sec:baralg}

A generative model $\mathbb{P}_{\theta}$ 
is a probability measure in $\mathcal{M}_{1}^{+}(\mathcal{X})$, parametrized by a vector $\theta$. Generative models are typically defined as push-forwards of a latent measure $\rho  \in \mathcal{M}_{1}^{+}(\mathcal{Z})$ on a lower-dimensional space through a generator function $G_{\theta}:\mathcal{Z}\rightarrow \mathcal{X}$. This means that a sample $\v{x}$ from $\mathbb{P}_{\theta}$ is obtained by first sampling $\v{z}$ from the latent $\rho$, then mapping it through $G_{\theta}$, i.e. $\v{x} = G_{\theta}(\v{z})$. More concisely, we simply write $\mathbb{P}_{\theta}=G_{\theta\#}\rho$.

In the context of estimating barycenters of measures, we propose to parametrize the barycenter using a generative model $\mathbb{P}_{\theta}$. This turns the problem of estimating the barycenter into finding optimal model parameters
\begin{align}\label{eq:approxbary}\theta^\star &= \argmin_{\theta} L(\theta), \\  L(\theta)& \coloneqq \sum_{p=1}^P\beta_p l_p(\theta),\qquad
    l_p(\theta)\coloneqq D(G_{\theta \#}\rho, \mu_{p}),
\end{align}
 where $D$ is a discrepancy between measures. Equation \eqref{eq:approxbary} (globally) parametrizes the barycentric problem \eqref{eq:bary} and can be solved by stochastic gradient descent as described in Algorithm \ref{algo:alg}. 
 
 In each training iteration, the algorithm receives a batch of data points from the individual measures as well as a batch of samples from the generator. Those are then used to compute  stochastic gradients $g_p(\theta)$ of the distances between the generator and each of the measures $\mu_p$. The model parameters $\theta$ are then updated by running gradient descent steps using the stochastic barycentric gradient $\sum_{p=1}^P\beta_p g_p(\theta)$.
We note that the discrepancy $D$ needs to be well-defined for measures with discrete support as the barycenter is only accessible through its samples.

\begin{algorithm}[ht!]
\centering
\begin{algorithmic} 
\REQUIRE Network $G_{\theta}$, measures $\{\mu_{p}\}_{p=1}^{P}$, weights $\{\beta_{p}\}_{p=1}^{P}$, base measure $\rho$, distances $\{D_{p}\}_{p=1}^{P}$, learning rate $\gamma$
\FOR{epoch in epochs}
\FOR{$p=1,..., P$}
\STATE Sample minibatches $\{x^{(p)}_{j} \sim \mu_{p} \}_{j=1}^{J}$
\STATE Sample $z_{j}\sim \rho$, $j=1,...,J$
\STATE Compute \begin{align*}g_p(\theta)=\nabla_\theta D_{p}(\sum\nolimits_{j=1}^{J}\delta_{\v x^{(p)}_{j}}, \sum\nolimits_{j=1}^{J}\delta_{G_{\theta}(\v z_{j})}  )\end{align*} 
\ENDFOR
\STATE Update $\theta = \theta - \gamma \sum_{p=1}^{P}\beta_{p} g_{p}(\theta) $
\ENDFOR
\end{algorithmic}
\caption{Algorithm for computing barycenters of arbitrary measures}
\label{algo:alg}
\end{algorithm}
\paragraph{Inductive Biases}  We can incorporate prior knowledge on the form of the barycenter  through the generator's structure (e.g., CNNs for barycenters of images) and leverage global basis functions (neural networks in particular). This enables scaling to high-dimensional settings, unlike Dirac-based approaches that suffer from the curse of dimensionality as they optimize locations of particles in a high-dimensional space. Note that the generator $G_\theta$ is not restricted to being a neural network, and domain knowledge can enable more efficient learning. For instance, if we know that the actual barycenter is Gaussian, we can set $\rho=\mathcal{N}(\m{0},\m{I})$, $G_{\theta}(\v{z}_n)=\m{S}^{\frac{1}{2}}\v{z}_n+\v{m}$ and optimize the mean $\v{m}$ and covariance $\m{S}$ using our algorithm as shown empirically in Section \ref{sec:experiments}. 

\paragraph{Optimization}
As discussed in Section \ref{sec:mmd}, MMD with fixed kernels is not a sensible metric on high-dimensional spaces. MMD with deep kernels (SMMD) alleviates this issue by defining a metric between measures over learned features. In that case, the kernel and the generator $G_\theta$ are trained adversarially, similar to \cite{NIPS2018_7904}.
Analogous adversarial formulations of these discrepancies were advocated for $\c{W}_\epsilon$ and $\c{S}\c{W}_\epsilon$ \citep{pmlr-v84-genevay18a,bunne2019gwgan}. All these approaches require  careful regularization of the critic (e.g., by penalizing its gradient \citep{NIPS2017_7159,binkowski2018demystifying,NIPS2018_7904} or weight clipping \citep{pmlr-v70-arjovsky17a}).

\paragraph{Special Cases}
The special case of computing the barycenter of a single measure ($P=1$) corresponds to the traditional implicit generative modeling objective. In that setting, different kinds of discrepancies $D$ have been considered, including MMD \citep{DBLP:conf/uai/DziugaiteRG15,NIPS2017_6815}, 1-Wasserstein \citep{pmlr-v70-arjovsky17a,NIPS2017_7159}, Sinkhorn divergence \citep{pmlr-v84-genevay18a}, and $\c{G}\c{W}_\epsilon$ \citep{bunne2019gwgan}. 

From a purely computational perspective, \citet{ijcai2019-483} train a Wasserstein GAN on a single dataset by randomly splitting that dataset into $P$ subsets and minimizing the average 1-Wasserstein between samples from the GAN and from those subsets. This is a special case in which the individual measures are all equal to the same data distribution. This implies that the barycenter coincides with such data distribution leading to a significantly simpler problem. 

In the case where all measures are Gaussians, \citet{rigolletbures} derive  the gradients of the Wasserstein barycenter functional with respect to the mean and variance of the barycenter and use SGD to learn it.

\begin{remark}
In the MMD case, the barycenter computed using our algorithm targets the mixture of the datasets. Note that the generative MMD barycenter could thus be estimated by training a normal MMD GAN ($P=1$) on the mixture of the datasets. However, training with the barycentric objective allows for larger batches per mode as training scales as $O(PN^2)$, where $N$ is the number of samples per mode and $P$ is the number of modes, instead of $O(P^2N^2)$ for GANs. 
\end{remark}
\subsection{Convergence Analysis}
\label{sec:barcon}

The non-convexity of the loss \eqref{eq:approxbary} with respect to  model parameters $\theta$ makes it hard to guarantee global convergence. However, we study local convergence to stationary points, which is challenging on its own since the divergence $D$ often results from an optimization procedure. Recently, \citet{10.5555/3327757.3327812} provided related results for the regularized Wasserstein distance. 

However, their approach cannot be applied to MMD and SMMD. We hence leverage different techniques to prove convergence for these discrepancies. More generally, we show that local convergence holds for all discrepancies that are either Lipschitz-smooth or weakly convex and Lipschitz-continuous, of which both entropic-regularized Wasserstein, Sinkhorn divergence  MMD, and scaled MMD are special cases.

\subsubsection{Smoothness}
Typical local convergence results rely on notions of smoothness. 
 \textit{Lipschitz smoothness} is the most commonly-used notion to guarantee local convergence.
\begin{definition}\label{def:l_smooth}
	A function  $L:\Theta \mapsto \mathbb{R} $ is $M$-Lipschitz smooth if there exists an $M\geq 0$, such that
	\begin{align}
    \Vert\nabla L(\theta)- \nabla L(\theta')\Vert\leq M\Vert\theta-\theta'\Vert\ \ \forall \theta,\theta'\in \Theta.
\end{align}
\end{definition}
Lipschitz smoothness of a function $L$ requires that the gradient of $L$ exists and is Lipschitz continuous.

\citet{10.5555/3327757.3327812} showed that entropic-regularized Wasserstein GANs (\eqref{eq:approxbary} with $P=1$ and $D=\c{W}_\epsilon$) is  $M$-Lipschitz smooth with respect to the generator parameters $\theta$. This is easily  extended to $\c{SW}_{\epsilon}$, and to the barycentric case ($P \geq 1$):
\begin{proposition}
 \label{prop:l_smoothness}
	Let $\mathcal{X}$ and $\mathcal{Z}$ be compact and $G_\theta$ Lipschitz and Lipschitz-smooth. Then, the  barycenter objective 
\begin{align}	
L(\theta) := \sum_{p=1}^P \beta_p (\c{S})\c{W}_{\epsilon}( G_{\theta\#}\rho,\mu_p)
\end{align}
 is $M$-Lipschitz smooth for $M \in \mathbb{R}^+$.
 
 Proof in Appendix  \ref{sec:proofthm_l_smooth}.
\end{proposition}

In the case of the optimized MMD, the discriminator is also learned leading to a non-concave problem. Therefore, the approach by \citet{10.5555/3327757.3327812} cannot be applied to guarantee $M$-Lipschitz smoothness of the resulting objective with respect to generator parameters $\theta$. We use a different approach 
that relies on the weaker notion of \emph{weak convexity}.
\begin{definition}\label{def:weak_convex}
	A function  $L:\Theta \to \mathbb{R} $ is  $C$-weakly convex if there exists a positive constant $C$, such that $ L(\theta) + C\Vert\theta \Vert^2$ is convex.
\end{definition}
The next result shows that (optimized) MMD is Lipschitz continuous and weakly convex, which will turn out to be sufficient to guarantee local convergence:
\begin{proposition}
\label{prop:l_convex_mmd}
Assume the kernel $k$ is Lipschitz and Lipschitz-smooth and functions $f_\psi \in \mathcal{E}$ are Lipschitz, Lipschitz-smooth, and 
absolutely continuous with respect to the parameters $\psi$ and inputs $\v{x}$. Further assume $G_\theta$ is Lipschitz and Lipschitz-smooth in $\theta$. Then,
\begin{align}
	L(\theta) := \sum_{p=1}^P \beta_p \mathrm{(S)MMD}^2( G_{\theta\#}\rho,\mu_p)
\end{align}
is weakly convex and Lipschitz.

Proof in Appendix \ref{sec:proofthm_l_smooth}.
\end{proposition}

Proposition \ref{prop:l_convex_mmd} states that the optimized MMD (and hence the barycentric objective) is \textit{weakly convex} and Lipschitz provided that the discriminator satisfies additional smoothness constraints. This is also useful in the case $P=1$ as it proves that several instantiations of MMD GANs \citep{binkowski2018demystifying,NIPS2018_7904} are also weakly convex (guaranteeing convergence; see Section \ref{sec:localconv}). Next, we show that local convergence holds in both cases.

\subsubsection{Local Convergence}
\label{sec:localconv}
When Lipschitz smoothness holds (as in Proposition \ref{prop:l_smoothness}), standard arguments guarantee convergence to a local stationary value $\theta^{\star}$ for gradient descent or SGD. When only  weak convexity and Lipschitz continuity hold  (as in Proposition \ref{prop:l_convex_mmd}), it is still possible to guarantee local convergence as shown in \citet{Davis:2018}. 
However, both cases require access to an unbiased estimate of the gradient of $L$.
In practice, this is not possible as $L$ is estimated by approximately solving an optimization problem. 

Therefore, we propose to use a similar setting as in \citep{10.5555/3327757.3327812}, where we assume access to an unbiased estimate of a direction $g$ that approximates $ \nabla L(\theta) $ to a precision $\delta$. In other words,  $g$ satisfies
$
\Vert\nabla L(\theta) - g\Vert^2\leq \delta^2,
$
and $\Tilde{g}$ is an  unbiased stochastic estimator  of $g$, i.e. $\mathbb{E}[\Tilde{g}]=g$, which we assume we have access to.
Such an estimate can be obtained by performing a few steps of gradient descent on the discriminator, in the case of the  SMMD, and then evaluating the gradient of the resulting loss with respect to $\theta$ on new samples. 
 We further assume that the noise in $\Tilde{g}$ has a bounded variance, i.e.   
 $\mathbb{E}[\Vert g -  \Tilde{g}\Vert^2]\leq \sigma^2$, and we define $\Delta := L(\theta_0)- \inf_{\theta} L(\theta)$ as the initial regret.  
 \begin{figure*}[ht!]
\centering
\subfigure[Nested ellipse]{
  \includegraphics[width=0.23\hsize]{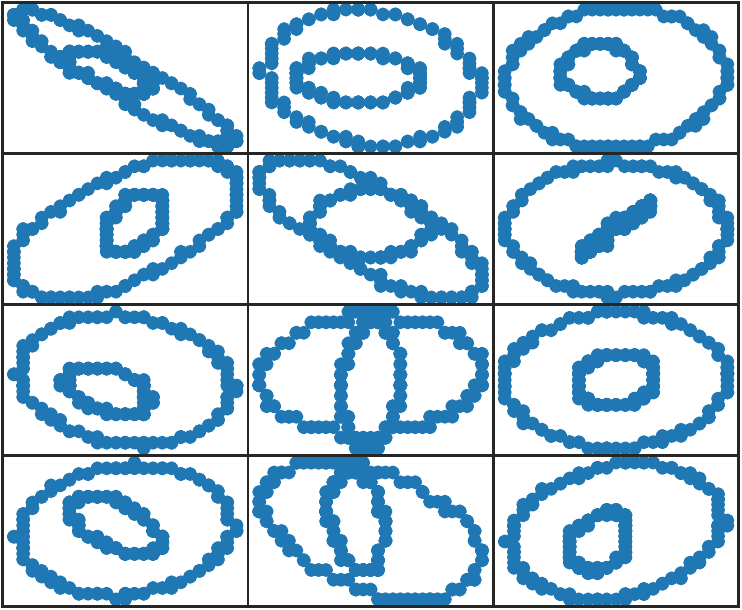}
  \label{fig:dataset_ell}
  }%
\subfigure[MLP]{
  \includegraphics[width=0.23\hsize]{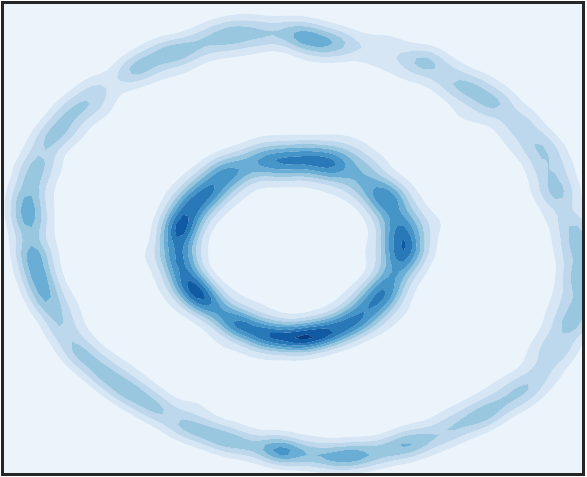}
  \label{fig:mlp}
  }
\subfigure[Structural model]{
  \includegraphics[width=0.23\hsize]{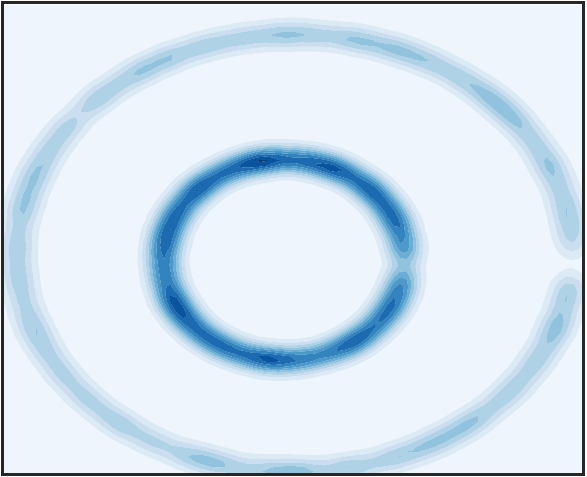}
  \label{fig:elli}
  }%
  \subfigure[\citet{NIPS2019_9130}]{
  \includegraphics[width=0.23\hsize]{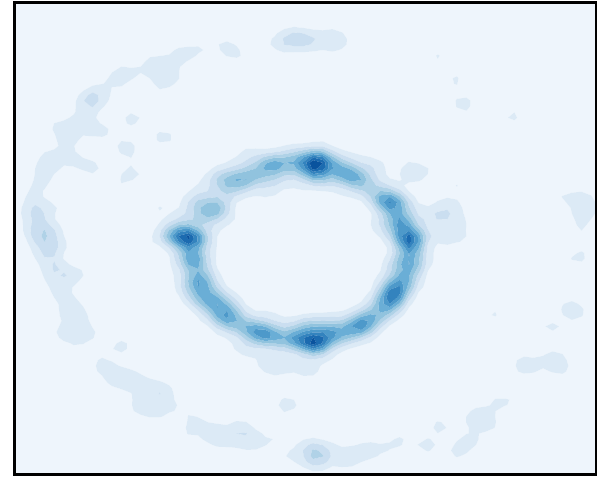}
  \label{fig:giulia}
  }%
\caption{Sinkhorn barycenter of 30 nested ellipses, of which a subset is displayed in \subref{fig:dataset_ell} using \subref{fig:mlp} the MLP parametrization, \subref{fig:elli} the nested ellipses parametrization, \subref{fig:giulia} \citet{NIPS2019_9130}.}
\label{fig:ellipses}
\end{figure*}
 \begin{figure*}[ht!]
\centering
\subfigure[Sinkhorn barycenter]{
  \includegraphics[width=0.30\hsize]{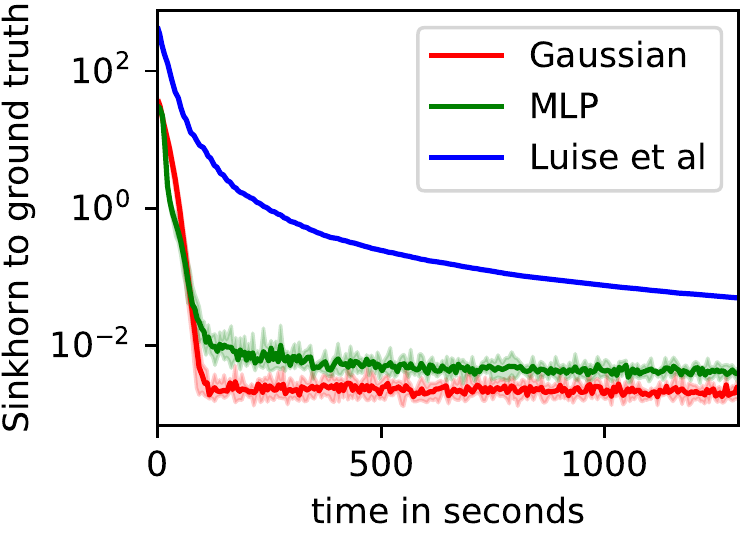}
  \label{fig:sink_gauss}
  }%
  \hfill
  \subfigure[Sinkhorn barycenter -- High Dim.]{
  \includegraphics[width=0.31\hsize]{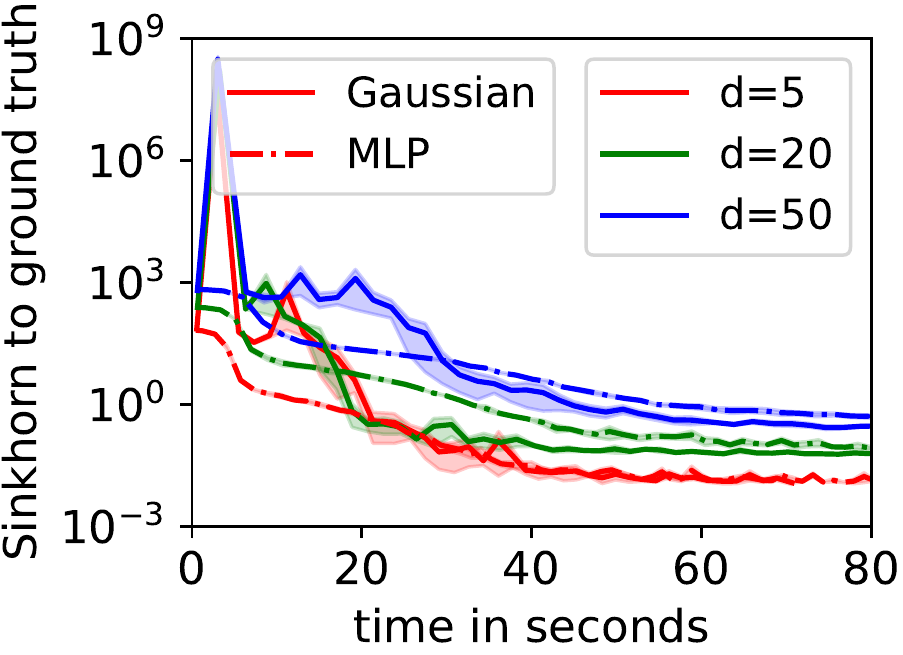}
  \label{fig:highdim}
  }%
  \hfill
  \subfigure[MMD barycenter -- High Dim.]{
  \includegraphics[width=0.30\hsize]{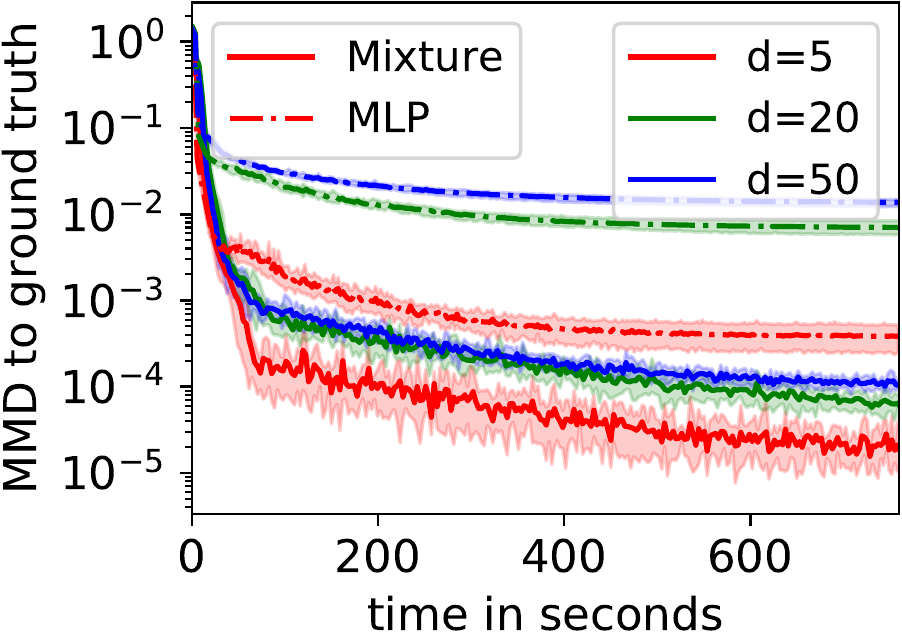}
  \label{fig:mmd_gauss}
  }
  
\caption{\subref{fig:sink_gauss}:Convergence plot of the computation of the barycenter of 15 Gaussians ($d=2$)  w.r.t.  $\c{S}\c{W}_\epsilon$ using MLP/Gaussian parametrizations, and \citet{NIPS2019_9130}. We also consider higher dimensions ($d=5,20,50$) in  \subref{fig:highdim} w.r.t $\c{SW}_\epsilon$ via MLP and Gaussian parametrizations, and in \subref{fig:mmd_gauss} w.r.t. MMD via a mixture parametrization.}
\label{fig:gaussians}
\end{figure*}
\begin{theorem}[\citep{10.5555/3327757.3327812}]
\label{thm:local_convergence}
   Assume $\Vert\nabla L(\theta) - g\Vert^2\leq \delta^2$, $\mathbb{E}[\Vert g -  \Tilde{g}\Vert^2]\leq \sigma^2$ and $\mathbb{E}[\Tilde{g}]= g$. 
  Also, if $L(\theta)$ is $M$-Lipschitz smooth (as in Proposition \ref{prop:l_smoothness}), then setting the learning rate to $\alpha:=\sqrt{\frac{2\Delta}{M\sigma^2}}$ yields \begin{equation}
        \min_{0\leq t\leq T-1} \mathbb{E}[||\nabla L(\theta_{t})||^{2}] \leq\sqrt{\frac{8\Delta M\sigma^2}{T}}+\delta^2.
        \label{eq:conv}
    \end{equation}
    \label{the:local}
\end{theorem}
Theorem \ref{thm:local_convergence}  shows that stochastic gradient methods converge to a stationary point 
when Proposition~\ref{prop:l_smoothness} holds.
If $L$ is only $C$-weakly convex as in Proposition \ref{prop:l_convex_mmd}, local convergence still holds \citep{Davis:2018}.

\section{Experiments}\label{sec:experiments}
We  demonstrate that our approach can scale the computation of barycenters to high dimensions, while still recovering accurate barycenters. We provide extensive experimental details in the Appendix.

We emphasize that, while MMD barycenters are known in closed form (mixture of measures), and that potentially simpler optimization schemes targeting it exist (GAN on the mixture of the datasets),   studying them empirically allows us to analyze the performance our algorithm. We also study barycenters for which a general closed form is not known, including Sinkhorn and SMMD barycenters. In such cases, a scalable algorithm is required, especially in high dimensions.

\subsection{Traditional Barycentric Problems}
We start with classical barycenter problems to demonstrate our approach yields sensible solutions to the barycentric problem \eqref{eq:bary}, and that leveraging structure can speed up computations.

\paragraph{Nested ellipses} We consider the computation of the $\c{S}\c{W}_\epsilon$ barycenter of $P=30$ nested ellipses, reproducing the example of \citet{pmlr-v32-cuturi14,NIPS2018_7827,NIPS2019_9130}. We compare to the algorithm proposed by \citet{NIPS2019_9130}. We consider two approaches to parametrizing the generator $G_\theta$, $(i)$ using a multi-layer perceptron (MLP) as $G_\theta$ and $(ii)$ exploiting inductive biases by parametrizing two ellipses ($\theta$: axis lengths and centers of both ellipses).  Figure \ref{fig:ellipses} shows that both approaches recover the barycenter, and obtain a similar but more accurate solution than the approach proposed in \citep{NIPS2019_9130} (under a time budget). In particular, there is  significantly more support on the ground truth barycenter due to the global nature of our algorithm.

\paragraph{Gaussians}
To illustrate the importance of the structural knowledge, we consider two different generative models for the barycenter: A model which contains the ground-truth barycenter (GT model) and a generic MLP network which doesn't explicitly encode structural knowledge about the barycenter. In the case of the MMD, the GT model is simply a mixture of Gaussians parametrized by their means and variances, while for $\c{S}\c{W}_\epsilon$, the GT model is  given by a single Gaussian \citep{janati}.
%
%

Figure \ref{fig:sink_gauss} shows that (i) our algorithm converges to a stationary point (\cref{sec:localconv}) and the gradient bias is negligible; (ii)  structural knowledge can lead to faster and more accurate approximations as the Gaussian parametrization converges to a better solution than the MLP; (iii) our algorithm is significantly faster than \citet{NIPS2019_9130} (runtimes/implementations discussion in Appendix). 
\begin{figure}[tb]
\centering
\subfigure[Sinkhorn Barycenter]{\includegraphics[width=0.48\hsize]{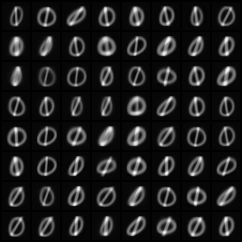}\label{fig:sink01}}
\subfigure[MMD Barycenter]{\includegraphics[width=0.48\hsize]{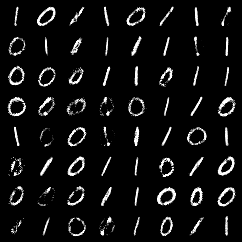}\label{fig:mmd01}}
\caption{Samples of digits from barycenters on MNIST datasets  of $0$s and $1$s with respect to Sinkhorn  \subref{fig:sink01} and MMD \subref{fig:mmd01}. We observe the  interpolation and mixture behaviors (See Propositions \ref{the:wass}, \ref{the:mmd}). We include barycenters of $0$s, $1$s, and $2$s in the Appendix.}
 \label{fig:sinkmnist}
\end{figure} 
Figures \ref{fig:highdim},\ref{fig:mmd_gauss} compare the GT model to the MLP model in higher dimensions for both $\c{SW}_\epsilon$ and MMD. In the case of $\c{SW}_\epsilon$ \subref{fig:highdim}, we observe that both GT model and MLP model recover accurate solutions of the barycentric  problem even in higher dimensions where the algorithm from \citet{NIPS2019_9130} does not apply ($d>2$). 
In the case of MMD (Figure \ref{fig:mmd_gauss}), the GT model outperforms the MLP model significantly, suggesting that an MLP is not necessarily a good model for mixtures of distributions. 

This is consistent with the discussion in \citet[Section 6.2]{DBLP:conf/birthday/BottouALO17}  which implies that implicit models families, such as MLPs, are better suited for parametrizing Wasserstein barycenters than MMD barycenters. 
We thus conclude that enforcing sensible inductive biases is essential to scaling to high dimensions.
\subsection{Barycenters of Natural Images} 
In the following, we demonstrate that the combination of structural knowledge and parametric models can scale barycentric computations to high dimensions. Previous papers considered problems in which measures are supported on low-dimensional spaces. Even in experiments with images, these were considered as densities on a 2D space \citep{pmlr-v32-cuturi14,NIPS2019_9130}. In the following, we consider a more challenging setting in which each measure consists of a dataset of $10^4$--$10^5$ images of dimension $10^3$--$10^5$. 

{\bf MNIST}  We define $\mu_m$ as the dataset of all $m^{th}$ MNIST digits (e.g. $\mu_0$ corresponds to the dataset of all MNIST $0s$). Each measure consists of approximately $5,000$ samples in a $32\times32$-dimensional space. 

We compute the Sinkhorn barycenter of $\mu_0, \mu_1$ in Figure~\ref{fig:sinkmnist} (left)  and of $\mu_0, \mu_1, \mu_2$ (Appendix). We use a moderate entropic coefficient;  hence, barycentric properties should be close to those of Wasserstein barycenters described in Proposition \ref{the:wass}. Both figures show the expected interpolation behavior, i.e., each sample from the barycenter is the interpolation of a `similar' 0 and 1 (Figure \ref{fig:sinkmnist} (Left)), and  of a `similar'  0, 1 and 2 (See in Appendix). 
Behaviors for barycenters of measures on Euclidean spaces (Figure \ref{fig:ellipses}) and on image spaces (Figure~\ref{fig:sinkmnist}) may at first seem contradictory. However, this is due to the fact that in the former case, a single atom of a specific measure consists of a point on an ellipse, whilst in the latter case it consists of a single image. Hence, interpolation on these two spaces is different as in the former case the overall barycenter will result in a smoothed out ellipse, whilst in the latter case it will result in a collection of interpolated (similar) images from the different classes.
We also compute the MMD barycenter of $\mu_0$ and $\mu_1$ (without optimized features) using our algorithm, which is expected to be a mixture of the datasets (see Proposition \ref{the:mmd}). Figure~\ref{fig:sinkmnist} illustrates the expected mixture behavior, which is in stark contrast to the interpolation behavior of Wasserstein barycenters.

\begin{figure}
\centering
\includegraphics[width=1.0\hsize]{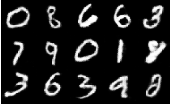}
\caption{Samples from our SMMD barycenter on MNIST datasets of different digits. We compute the barycenter of $\mu_0,.., \mu_9$ ($\mu_i$ is the dataset of all $i^{th}$ digit).} \label{fig:mnist}
\end{figure}

\begin{figure}
\centering
    \includegraphics[width=1.0\linewidth]{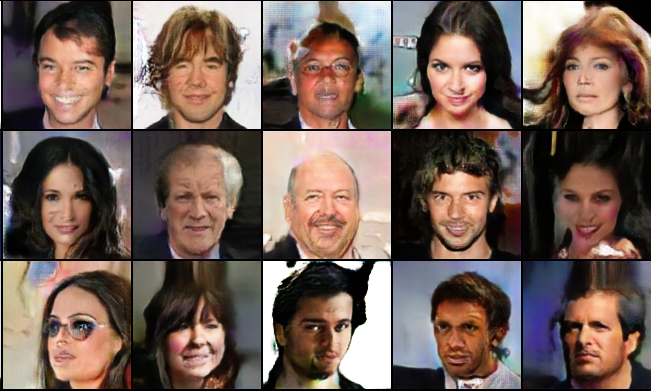}
\caption{Samples from our SMMD barycenter of all CelebA females and males (respectively $\mu_0$ and $\mu_1$).}    \label{fig:celeba}
\end{figure}

To continue, we compute the SMMD (MMD with optimized features) barycenter of 10 measures $\mu_0,...,\mu_{9}$ and emphasize that SMMD barycentric properties are not known in closed form. Figure~\ref{fig:mnist} shows that the SMMD barycenter generates meaningful samples from all classes. Its behavior is similar to the mixture behavior of MMD barycenters (see Proposition \ref{the:mmd}). In that case, barycenters average measures over features instead of over images themselves, which is in contrast to the MMD barycenter computed in Figure~\ref{fig:sinkmnist}.
\paragraph{CelebA} We finally compute the SMMD barycenter of two measures, CelebA males and females, each having approximately $100,000$ locations (images). Images are re-scaled to  $3\times 128\times 128$ pixels, so that each (males/females) lives in an approximately $50,000$-dimensional space. We use deep convolutional generators and critics to leverage the structural knowledge about the input locations (images).
Figure \ref{fig:celeba} illustrates that the SMMD barycenter generates meaningful high-quality samples from both measures. Overall, $(i)$ expected barycentric geometric properties are observed in high-dimensional problems;  $(ii)$ using structural knowledge (here a CNN) enables (good) approximate solutions to barycentric problems at unprecedented scale.

\section{Discussion}

Our proposed approach relies on global parametric structured models and thus departs significantly from previous barycentric works with local unstructured models \citep{pmlr-v32-cuturi14,NIPS2017_6858,NIPS2018_8274,pmlr-v32-cuturi14,pmlr-v80-claici18a,NIPS2019_9130}. This allows us to scale to higher dimensions under the assumption that inductive biases on the optimal solution are known. Such biases can be enforced through the structure of the parametric model (e.g., CNNs for SMMD barycenters of measures over images). 
Without enforcing structure, barycentric algorithms are doomed in high dimensions as studied by \citet{altschuler}.
However, we note that in low-dimensional problems, where absolutely no structure is known about the barycenter, more brute-force approaches that do not enforce inductive biases (e.g., \citep{NIPS2019_9130}) may be more appropriate.

Our approach also departs from classical GAN problems (recovering them in the case $P=1$). Indeed, we aim to find a model that achieves the best trade-off between multiple distribution according to some distance. Hence, the choice of the distance has a significant impact on the nature of the solution. This is unlike GANs where the goal is to approximate the data distribution and where the choice of the distance has little impact on the nature of the optimal solution \cite{lucic2017gans}. 
 Our work can hence be considered a generalization of their works in two orthogonal directions:  i) the averaging direction (we consider $P>1$ measures), and ii) the distance direction as we consider general choices of discrepancies between measures. 
Finally, we provided local convergence guarantees instead of global ones 
due to the non-convexity of the objective. While \citet{NIPS2019_9130} provided global convergence guarantees, they only hold under the assumption that an inner non-convex problem is solved exactly. In general, this problem remains as challenging as ours.

\section{Conclusion}
We proposed an algorithm for estimating high-dimensional barycenters of probability measures with respect to general choices of discrepancies. The key idea is to leverage a different parametrization of the barycenter. This turns the barycentric problem into a problem of learning model parameters, thereby sidestepping the curse of dimensionality from which other algorithms for estimating barycenters suffer. Our approach also enables incorporating explicit structural inductive biases  in the model (e.g., CNNs for measures over images). We proved local convergence of our algorithm to stationary points under mild smoothness assumptions on the discrepancy considered. 
We applied our algorithm to problems at an unprecedented scale (for both Sinkhorn and SMMD discrepancies), which includes estimating barycenters of measures with more than $10^5$ locations in over $10^4$ dimensions.

\section*{Acknowledgments}
We are grateful to Giulia Luise for providing us code and data for experiments, and for providing feedback on the draft. 
SC was supported by the Engineering and Physical Sciences Research Council (grant number EP/S021566/1).

\bibliographystyle{abbrvnat}
\bibliography{references}

\onecolumn
\appendix

\section{Proof of    Propositions \ref{prop:l_smoothness} and   \ref{prop:l_convex_mmd}}
\label{sec:proofthm_l_smooth}

We start by introducing some key notation. We denote by $\mathcal{E}$ the set of discriminators $f$ in the optimized MMD
\begin{align}\label{eq:optimized_mmd}
	SMMD^2(G_{\theta},\mu) := \sup_{f\in \mathcal{E}}MMD_f^2(G_{\theta\#}\rho,\nu).
\end{align}
We next state the assumptions that will be used in the following.
\begin{enumerate}
	\item\label{assump:compactness} $\mathcal{E}$ is parametrized by a compact set of parameters $\Psi$ and any $f\in \mathcal{E}$ is continuous w.r.t. those parameters. 
	\item\label{assump:l_smooth_critic} Functions in $\mathcal{E}$ are jointly continuous w.r.t. $(\psi,\v{x})$  and are $L$-Lipschitz and $L$-Lipschitz smooth w.r.t. to the input $\v{x}$, i.e.,
		\begin{align}
			\Vert f_{\psi}(\v x)-f_{\psi}(\v x') \Vert &\leq L\Vert \v x-\v x' \Vert,\\
			\Vert \nabla_{\v{x}} f_{\psi}(\v x)-\nabla_{\v{x}} f_{\psi}(\v x') \Vert&\leq L\Vert \v x-\v x' \Vert.
		\end{align}
	\item\label{assump:l_smooth_gen} There exists a square integrable function $a :\mathcal{Z}\to \mathbb{R} $ and an integrable function $b:\mathcal{Z}\to \mathbb{R}$, such that generators $\theta \mapsto G_{\theta}(z)$ are $a$-Lipschitz and $b$-Lipschitz smooth in the following sense:
			\begin{align}
				\Vert G_{\theta}(\v z)-G_{\theta'}(\v z) \Vert&\leq \vert a(\v z)\vert \Vert \theta - \theta' \Vert,\\
				\Vert \nabla_{\theta} G_{\theta}(\v z) - \nabla_{\theta} G_{\theta'}(\v z)  \Vert &\leq \vert b(\v z) \vert\Vert  \theta -\theta' \Vert.  
			\end{align}
			Moreover, for all $ h\in \mathcal{E}$ and $\theta \in \Theta$ the square integral of $a$ and integral of $b$ are uniformly bounded by some constant $C$ so that
			\begin{align}
				\int \vert a(\v z)\vert^2 p_{h,\theta}\circ g_{\theta} d\eta \leq C,\\
				\int \vert b(\v z)\vert p_{h,\theta}\circ g_{\theta} d\eta \leq C.
			\end{align}
		\item \label{assump:sup_gen_smooth} $G_{\theta}$ is $L$-Lipschitz and $L$-Lipschitz smooth in $\theta$ uniformly in $\v z$.
		\item \label{assump:compact_space} The input and output spaces are compact.
		\item \label{assump:linear_growth_kernel} The kernel is $L$-smooth and $L$-Lipschitz.
\end{enumerate}

\begin{proposition}\label{prop:MMD_l_smooth}
	Under assumptions \ref{assump:l_smooth_critic}, \ref{assump:l_smooth_gen} and \ref{assump:linear_growth_kernel}, we have that $ \mathcal{M}(\theta) := MMD^2_{f}(G_{\theta\#}\rho,\mu)$ is Lipschitz and Lipschitz smooth uniformly on $\mathcal{E}$, i.e.,
	\begin{align}
		\vert \mathcal{M}_{\theta}(f) -  \mathcal{M}_{\theta'}(f) \vert &\leq L \Vert \theta - \theta' \Vert\\
		  \Vert \nabla \mathcal{M}_{\theta}(f) -  \nabla\mathcal{M}_{\theta'}(f)\Vert &\leq   L \Vert \theta - \theta' \Vert.
	\end{align}
\end{proposition}
\textit{Proof:}
Recall that under assumptions \ref{assump:l_smooth_critic}, \ref{assump:l_smooth_gen} and \ref{assump:linear_growth_kernel}, the dominated convergence theorem guarantees that $\mathcal{M}_{\theta}(f)$ is differentiable in $\theta$, with a gradient
\begin{align}
	\nabla_{\theta} \mathcal{M}_{\theta}(f) &= 2(\int \nabla_1 k(f\circ G_{\theta}(\v z),f(\v x))R_{\theta}(\v z)d\mu(\v x)d\rho(\v z) \\&\quad -\int \nabla_1 k(f\circ G_{\theta}(\v z),f\circ G_{\theta}(\v z))R_{\theta}(\v z)d\mu(\v x)d\rho(\v z)),
\end{align}
where $  R_{\theta}(\v z) = \nabla f(G_{\theta}(\v z))\nabla_{\theta}G_{\theta}(\v z)$. Moreover, the gradient can be upper-bounded uniformly in $f$ since $f$, $G_{\theta}$ and $k$ are all Lipschitz. This implies that $\mathcal{\theta}(f)$ is Lipschitz uniformly in $f$. The fact that $\nabla \mathcal{M}_{\theta}(f)$ is uniformly Lipschitz also results the fact that gradients of  $f$, $G_{\theta}$ and $k$ are all Lipschitz.

\textit{Proof of  Proposition \ref{prop:l_smoothness}.} Here, we  use \citep{10.5555/3327757.3327812} (Theorem 3.1), which guarantees that the entropy-regularized Wasserstein distance is smooth as soon as Assumptions \ref{assump:sup_gen_smooth} and \ref{assump:compact_space} hold. This implies that the Sinkhorn divergence is also smooth and, finally, that $\theta \mapsto L(\theta)$ is smooth as a convex combination of smooth functions.

\textit{Proof of Proposition \ref{prop:l_convex_mmd}.} 
We will only prove that the optimized MMD is $L$-weakly convex. The resulting loss $L$ will then also be weakly convex for a suitable constant as it is simply a convex combination of weakly convex terms.  For simplicity, we denote $\mathcal{SM}_{\theta} := SMMD^2(G_{\theta\#}\rho,\mu)$.

Using Proposition \ref{prop:MMD_l_smooth}, we know that $\mathcal{M}(\theta) := MMD^2_{f}(G_{\theta\#}\rho,\mu)$ is $C$-smooth. It is therefore weakly convex and the following inequality holds:
\begin{align}
\mathcal{M}_{\theta_t}(f) \leq  t \mathcal{M}_{\theta}(f) + (1-t)\mathcal{M}_{\theta'}(f) + \frac{C}{2}t(1-t)\Vert\theta-\theta'\Vert^2.	
\end{align} 
Taking the supremum w.r.t. $f$, it follows that
\begin{align}
	\mathcal{SM}(\theta_t)\leq t \mathcal{SM}(\theta) + (1-t)\mathcal{SM}(\theta') + \frac{C}{2}t(1-t)\Vert\theta-\theta'\Vert^2.
\end{align}
This means exactly that $\mathcal{SM}$ is weakly convex.

The fact that $\mathcal{SM}$ is Lipschitz, is a consequence of Proposition \ref{prop:MMD_l_smooth}. Indeed, $\mathcal{M}_{\theta}(f)$ is Lipschitz  in $\theta$ uniformly on $\mathcal{E}$. Hence,
\begin{align}
	\mathcal{M}_{\theta}(f) \leq \mathcal{M}_{\theta}(f)  +C \Vert \theta - \theta' \Vert.
\end{align}
Taking the supremum over $f$, it follows directly that
\begin{align}
	\mathcal{SM}(\theta)\leq \mathcal{SM}(\theta') +C \Vert \theta - \theta' \Vert.
\end{align}
By exchanging the roles of $\theta$ and $\theta'$, we get the other side of the inequality. $\mathcal{SM}(\theta)$ is indeed Lipschitz in $\theta$ and by the Rademacher theorem, $\mathcal{SM}$ is even differentiable for almost all $\theta$.

\section{Proof of Wasserstein barycentric properties}
\label{proof:w2behave}

Here we consider the barycenter problem when the $\c{W}_2$ distance is used:
\begin{align}\label{eq:w_2_barycenter}
	\min_{P} L(P):=\sum_{k}\alpha_k \c{W}_2^2(P,P_k) 
\end{align}
We will show that the optimal $P$ exists and can be obtained by solving the multi-marginal problem
\begin{align}\label{eq:multi_marginal}
	\min_{Q}  \int  \sum_{k}\alpha_k \Vert \v x_k-T(\m X) \Vert^2 ~dQ(\m X)  ,
\end{align}
where $\v{X} = (\v{x}_1,...,\v{x}_P)$, $T(\v{X}) = \sum_k \alpha_k \v{x}_k $ and $Q$ is a coupling between $\v{x}_1,\dotsc,\v{x}_P$ with marginals given by $(P_k)_{1\leq k\leq P}$.
A key remark is that \eqref{eq:multi_marginal} is equivalent to 
\begin{align}\label{eq:multi_marginal2}
	\max_{Q}  \int  \Vert T(\v{X}) \Vert^2 ~dQ(\v{X}). 
\end{align}
This is simply a consequence of expanding the square in \eqref{eq:multi_marginal} and using the definition of $T(\v{X})$. We denote by $Q^{\star}$ the optimal solution for \eqref{eq:multi_marginal} for which we have by definition
\begin{align}\label{eq:inequality_multi_marginal}
	\int  \Vert T(\v{X}) \Vert^2 ~dQ^{\star}(\v{X})\geq \int  \Vert T(\v{X}) \Vert^2 ~dQ(\v{X})
\end{align} 
for all multi-marginal coupling $Q$ of $(P_k)_{1\leq k\leq P}$.

Consider now $P^{\star}= T_{\#}Q^{\star}$ where a sample $\v{Y}$ is obtained by first sampling $\v{X}$ according to $Q^{\star}$ and then setting $\v{Y}=T(\v{X})$. We obtain an upper bound on $L(P^{\star})$ via
\begin{align*}
	L(P^{\star}) &= \sum_k \alpha_k \c{W}_2^2(P^{\star},P_k)\\
	&\leq \sum_k \alpha_k \int \Vert T(\v{X})-\v{x}_k \Vert^2 ~dQ^\star(\v{X}) \\
	&= \int \left(\sum_{k}\alpha_k \Vert \v{x}_k \Vert^2 - \Vert T(\m{X}) \Vert^2\right) d Q^\star(\m{X})\\
	&=  \sum_{k}\alpha_k\int \Vert \v{x}_k \Vert^2 d P_k(\v{x}_k) - \int \Vert \m{Y} \Vert^2 dP^{\star}(\m{Y}).
\end{align*}
The second line is obtained by using the fact that $(T,Proj_k)_{\#}Q^\star$ defines a coupling between $P^{\star}$  and $P_k$. The third and last lines are expansions recalling  that the marginals of $Q^\star$ are $P_k$ and that $P^{\star}= T_{\#}Q$.

Now, let $P$ be any probability distribution with finite second moment. It is well known that there exist optimal couplings $\pi_k$ between $P$ and each $P_k$, such that 
\begin{align}
	\c{W}_2^2(P,P_k) = \int \Vert \m{Y}-\v{x}_k \Vert^2 d\pi_k(\m{Y},\v{x}_k).
\end{align}
Moreover, by Proposition \ref{lem:gluing}, we know there exists a joint coupling $\pi$ between $(\m{Y},\v{x}_1,...,\v{x}_P)$ with pairwise marginals given by $\pi_k$. Hence, $L(P)$ can be expressed as
\begin{align}
	L(P) &=\int  \sum_k \alpha_k \Vert \m{Y}-\v{x}_k \Vert^2 ~d\pi(\m{Y},\m{X})\\
	&= \int \left( \Vert \m{Y} \Vert^2 - 2 \m{Y}^{\top}T(\m{X}) + \sum_k \alpha_k\Vert \v{x}_k\Vert^2\right) d \pi(\m{Y},\m{X})\\
	&= \int \Vert \m{Y} \Vert^2 dP(\m{Y}) - 2\int \m{Y}^{\top}T(\m{X})\pi(\m{Y},\m{X}) + \sum_k \alpha_k \int \Vert \v{x}_k\Vert^2 d P_k(\v{x}_k).
\end{align}
The first line is by definition of the coupling $\pi$, the second line is a simple expansion of the square function and last line uses that $\pi$ has marginals given by $P$ and $(P_k)_{1\leq k\leq P}$.

Using the preceding expressions, we  now compute a lower bound on the difference $L(P)-L(P^{\star})$ as
\begin{align}
	L(P)-L(P^{\star}) \geq \int \Vert \m{Y} \Vert^2  dP(\m{Y}) - 2\int \m{Y}^{\top}T(\m{X})\pi(\m{Y},\m{X}) + \int \Vert \m{Y} \Vert^2 dP^{\star}(\m{Y}).
\end{align}
Consider now $Q_0$ be the distribution over $\m{X}$ obtained by marginalizing $\pi$ over $\m{Y}$. Then $\pi$ is a coupling between $P$ and $Q_0$. Moreover, by definition of $Q^{\star}$  we have that
\begin{align}
	\int \Vert \m{Y} \Vert^2 dP^{\star}(\m{Y}) = \int \Vert T(\m{X}) \Vert^2 dQ^{\star}(\m{X})\geq \int \Vert T(\m{X}) \Vert^2 dQ_0(\m{X}).
\end{align}
This directly implies that
\begin{align}
	L(P)-L(P^{\star})&\geq \int \Vert \m{Y} \Vert^2  dP(\m{Y}) - 2\int \m{Y}^{\top}T(\m{X})\pi(\m{Y},\m{X}) + \int \Vert T(\m{X}) \Vert^2 dQ_0(\m{X})\\&= \int \Vert \m{Y}-T(\m{X}) \Vert^2\pi(\m{Y},\m{X})\geq 0.
\end{align}

\begin{proposition}\label{lem:gluing} 
	Given pairwise couplings $\pi_k$ between variables $(\m{Y},\v{x}_k)$ for $1\leq k\leq P$, there exists a joint coupling $\pi$ between $(\m{Y},\v{x}_1,...,\v{x}_P)$ that admits $\pi_k$ as marginals. (see \citep[Remark 10.2]{47813} or \citep[Gluing lemma, p. 24]{villani}).
\end{proposition}

\section{Proof of MMD properties}
\label{sec:proofmmd}
Here, we consider the MMD with a fixed kernel $k$. Denote by $\eta(P)$ the kernel mean embedding of the distribution $P$, ie.:  $\eta(P) = \int k(\v x,.)~dP(\v x) $. We want to show that $P^\star=\sum_k\alpha_k P_k$ is the minimizer of 
\begin{align}\label{eq:mmd_bar}
     \min_{P} \sum_k \alpha_k \Vert \eta(P)-\eta(P_k)\Vert_{\mathcal{H}}^2.
\end{align}
This is equivalent to finding an optimal function $\Phi$ in $\mathcal{H}$ that minimizes 
\begin{align}\label{eq:mmd_bar_rkhs}
     \min_{\Phi} \sum_k \alpha_k \Vert \Phi-\eta(P_k)\Vert_{\mathcal{H}}^2.
\end{align}
under the additional constraint that $\Phi$ is a mean embedding of some probability distribution $P$.
We will show that the unconstrained problem in \eqref{eq:mmd_bar_rkhs} admits $\eta(P^\star)$ as an optimal solution. Equation \eqref{eq:mmd_bar_rkhs} is a strongly convex quadratic function of $\Phi$. Therefore it admits a unique global minimum, which is given by the first-order optimality condition
\begin{align}
    \phi^{\star} = \sum_{k}\alpha_k \eta(P_k).
\end{align}
Now we use the fact that the kernel mean embedding is a linear operator on measures, which implies directly that $\sum_{k}\alpha_k \eta(P_k) = \eta(\sum_{k}\alpha_k P_k )=\eta(P^{\star})$.  We have shown that $\Phi^{\star}$, the unconstrained solution of \eqref{eq:mmd_bar_rkhs}, is a mean embedding for $P^{\star}$. This directly implies that $P^{\star}$ is an optimal solution to \eqref{eq:mmd_bar}. Uniqueness is obtained whenever the mean embedding is injective, i.e., the kernel $k$ is characteristic.

\section{Experimental Details}

\subsection{Nested Ellipses}
\subsubsection{Setup}
We compute the Sinkhorn divergence using Geomloss. For both parametrizations, we train using the Sinkhorn divergence with entropic coefficient $\epsilon=0.1$ and a batch size of 150.
\textbf{MLP parametrization} We use a MLP with 4-hidden layers (50, 200, 1000, 200 neurons), ReLU activations, a latent dimension of 10.  
\textbf{Ellipse parametrization}
We initialize the centers and axis of the nested ellipses from standard Gaussians.
\subsubsection{Discussion}
We note that if given a substantially higher time budget, the algorithm of \citet{NIPS2019_9130} would converge to a significantly better solution, as per its convergence guarantees. However, because of computational time constraints, we fixed the maximum number of support points to be added to $N=1500$, which resulted in the provided figure. By contrast, our approach leverages global basis functions, which in turn put mass on a large support directly, without having to optimize locations individually. 
\subsection{Gaussians}
\subsubsection{Setup}
We plot mean and the 5\%--95\% quantiles (across 5 random seeds). We compute the Sinkhorn divergence using Geomloss. We average over 5 seeds, and use an exponential scheduler with decay parameter $\lambda=0.985$ (the learning rate decreases every epoch). 
\textbf{Sinkhorn: MLP parametrization} We use a MLP with 4-hidden layers (50, 200, 1000, 200 neurons), ReLU activations, a latent dimension of 2, and the batch size to 150. We set the learning rate to $8\times 10^{-4}$ (for the high dim. experiment we set the latent dimension to 5.)
\textbf{Sinkhorn: Gaussian parametrization} We parametrize the mean and the variance of an isotropic $2$D Gaussian. We set the learning rate to $0.4$ (low-dimensional) and $2\times 10^{-2}$ (high-dimensional) and the batch size to 150. 

\subsubsection{Discussion}
We set the learning rate of the MLP and the Gaussians parametrization to the maximum value at which optimization is stable. In turn, we could set the latter's learning rate to a significantly larger value than the former's.

\subsection{Natural Images}
 For the SMMD experiment, we use DCGAN-like architectures for both the generator and critics (we use a different critic for each measure). We use the formulation of \citet{binkowski2018demystifying}, in particular a mixture of rational quadratic kernel, convolutional critics, along with gradient penalty. For the CelebA experiment, we also use  and spectral normalization for regularization. However, we set the critics' output dimensions to $1$ instead of $16$, which leads to similar performance. We perform five critic iterations per generator iteration and train using the ADAM optimizer with $\beta_1=0.5$, $\beta_2=0.99$ and a learning rate of $2\times 10^{-4}$. In CelebA experiments, we include an exponentially decreasing scheduling ($\gamma=0.99$). 
 
 For the MMD and Sinkhorn experiments, we set the batch size to $300$, the learning rate to $2\times 10^{-4}$. For MMD, we use a rational quadratic kernel with lengthscale $l=2$. We use a MLP with 4-hidden layers (50, 200, 1000, 200 neurons) as generator, ReLU activations, a latent dimension of 10. We do not use critics.

\end{document}